\documentclass[letterpaper, 10pt, conference]{ieeeconf}

\IEEEoverridecommandlockouts
\usepackage{amsmath,amssymb}
\usepackage{graphicx}
\usepackage{booktabs}
\usepackage{array}
\usepackage[hidelinks]{hyperref}
\usepackage{url}
\usepackage{textcomp}
\usepackage{times}
\usepackage{cuted}    
\usepackage{capt-of}  

\makeatletter
\def\maxwidth{\ifdim\Gin@nat@width>\columnwidth\columnwidth\else\Gin@nat@width\fi}
\def\maxwidthwide{\ifdim\Gin@nat@width>\textwidth\textwidth\else\Gin@nat@width\fi}
\makeatother

\title{\LARGE \bf Contact-Stable Deformable Tissue Simulation Using Implicit Integration and Live-Pose Grasp Constraints for  Laparoscopic Surgery Robot Policy Evaluation}

\author{%
Juahn Oh*$^{1,2,10}$, Dongho Yee$^{1,2,4,5}$, Jinseok Lee$^{2,4}$, Jiyul Lee$^{1,2,3}$, Yechan Seo$^{1,2,3}$,\\
Seong Jeong$^{1,2,3}$, Minsung Kim$^{1,2,4}$, Seonho Shim$^{2,11}$, Younghoon Noh$^{2,4}$, Hyuk Choi$^{1,2,3}$,\\
Youngbin Kong$^{1,9}$ and Hyoun-Joong Kong†$^{1,3,8}$%
\thanks{*First author. †Corresponding authors. $^{1}$Department of Transdisciplinary Medicine, Seoul National University Hospital, Seoul, Republic of Korea. $^{2}$Rosota Inc., Seoul, Republic of Korea. $^{3}$Department of Medicine, Seoul National University College of Medicine, Seoul, Republic of Korea. $^{4}$Department of Mechanical Engineering, Seoul National University, Seoul, Republic of Korea. $^{5}$Department of Computer Science and Engineering, Seoul National University, Seoul, Republic of Korea. $^{6}$Department of Surgery, Seoul National University Hospital. $^{7}$Department of Surgery, Seoul National University College of Medicine. $^{8}$Institute of Convergence Medicine with Innovative Technology, Seoul National University Hospital, Seoul, Republic of Korea. $^{9}$Interdisciplinary Program in Medical Informatics, Seoul National University College of Medicine. $^{10}$Eulji University College of Medicine, Daejeon, Republic of Korea. $^{11}$Department of Mechanical Engineering, Chungang University, Seoul, Republic of Korea.}%
}

\begin{document}

\maketitle
\thispagestyle{empty}
\pagestyle{empty}
\begin{strip}
  \includegraphics[width=\textwidth,height=0.4\textheight,keepaspectratio]{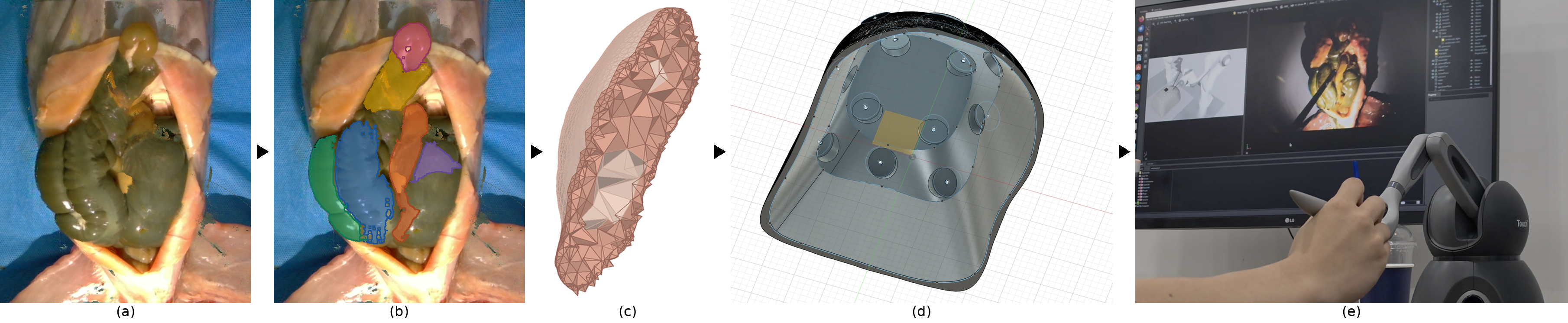}
  \captionof{figure}{From a fixed-view recording to a closed-loop surgical simulator.
    (a)~RGB-D recording with instruments removed by compositing.
    (b)~Organ segmentation.
    (c)~Tetrahedralised organ, cut away.
    (d)~Laparoscopic trainer with measured trocar ports.
    (e)~Bimanual cell under haptic teleoperation.
    (f)~In-vivo laparoscopy. Left: simulated endoscope view. Right: real endoscope frame.
    (g)~Left: simulated endoscope view of the bimanual cell. Right: real surgical cell with two FR3 arms through 6\,mm trocars.}
  \label{fig:overview}
\end{strip}

\begin{abstract}
Closed-loop evaluation of surgical robots requires tissue that deforms, can be grasped and lifted, and reproduces the anatomy in which the robot will operate. We present a simulator in which this tissue is reconstructed from a fixed-view RGB-D recording of the surgical field, composited to remove the instruments, closed into watertight volumes and tetrahedralised; the pipeline was applied unchanged to three specimens of two species (thirteen organs, 146,061 tetrahedra, no inverted elements). For one specimen, the organs are placed in a bimanual cell in which two Franka FR3 arms operate motorised instruments through 6~mm trocars. The core contribution is the numerical and contact design that keeps this cell stable: implicit integration, simulation meshes separate from collision meshes, numerical guards, and a grasp constraint captured at the live tissue pose. In 45 repeated grasp-lifts, a friction grasp held the tissue in 0 of 15 trials and each constraint grasp in 13 of 15; on displaced tissue, a rest-pose constraint produced one-step snaps of up to 17.8~mm, which live-pose capture eliminates. Against the recording, front-surface depth error is 1.33 to 1.41~mm, organ silhouette IoU is 0.80, and in five grasp-lifts reproduced from video the landmark displacement RMSE is 11.8~mm against 14.2~mm for a static prediction. Biofidelity is not claimed; the environment is intended for closed-loop feasibility, safety, contact and policy screening.
\end{abstract}

\section{Introduction}

Screening a surgical policy or a teleoperation interface before it is applied to an animal requires a closed-loop environment whose organs deform, can be grasped and lifted, and are arranged as in the field where demonstrations were recorded. Neither a hand-authored scene nor a CT-derived model reproduces this arrangement and appearance. We converted the recording already acquired during data collection, a RealSense D405 looking down at the surgical field, into simulable tissue.

Reconstruction was the smaller part of this work. A tetrahedral mesh extracted from a marching-cubes surface is not yet a soft body that two robot arms can grasp through trocars: at the element size of this anatomy, non-implicit integrators diverge, voxel resolution alters bending stiffness, a shared simulation and collision mesh leaves the organ static, and a naive grasp constraint displaces the instrument from its port. With these issues resolved, the environment measures grasp position, tissue lift, jaw engagement, RCM deviation, tracking error, contact force and stage progress. Fig.~1 summarises the route and Fig.~2 the architecture. The contributions are:

\begin{enumerate}
\item \textbf{Specimen-specific reconstruction.} A fixed-view surgical RGB-D recording is converted into watertight soft bodies, demonstrated on three specimens (\S{}III).
\item \textbf{Contact-stable robotic simulation.} Implicit integration, discretisation-aware modelling, separate simulation and collision meshes, numerical guards and a live-pose grasp constraint enable bimanual manipulation through trocars (\S{}IV--\S{}VI).
\item \textbf{Validation against measurement.} Depth error, a grasp-design ablation, real-to-sim similarity metrics, closed-loop RCM behaviour and throughput (\S{}VII).
\end{enumerate}

\section{Related Work}

\subsection{Surgical Simulators and Deformable Tissue Models}

Surgical simulation. Established frameworks provide interactive soft-tissue physics (SOFA~\cite{faure2012sofa}) and learning environments for dVRK-compatible~\cite{kazanzides2014open} and laparoscopic robots (SurRoL~\cite{xu2021surrol}, LapGym~\cite{scheikl2023lapgym}, ORBIT-Surgical~\cite{yu2024orbit}, Surgical Gym~\cite{schmidgall2024surgical}); they typically take the anatomy as an input asset. We do not propose a new framework but integrate specimen-specific reconstruction with contact-stable manipulation under trocar constraints.

\subsection{Real-to-Sim Reconstruction for Surgical Robotics}

Deformable tissue for manipulation. Position-based dynamics~\cite{muller2007pbd,macklin2016xpbd} is a standard choice for interactive soft tissue, including differentiable variants for trajectory optimisation~\cite{liang2024autopeel}. At our element size it did not remain bounded (\S{}VII-B); we use Vertex Block Descent~\cite{chen2024vbd}.

\subsection{Contact Modeling for Tissue Manipulation}

Real-to-sim for surgical tissue. Deformable tissue has been registered to position-based simulation from surgical perception~\cite{liu2021real,li2020super}, and residual mappings have been learned to reduce the remaining gap~\cite{liang2024residual}. Deformable neural reconstruction~\cite{wang2022neural} recovers endoscopic geometry and appearance for rendering; PhysTwin~\cite{jiang2025phystwin} builds simulable deformable objects from RGB-D video and evaluates them with Chamfer distance, tracking error and image similarity. We report comparable metrics for surgical tissue grasped by instruments. The screened policies are diffusion policies~\cite{chi2023diffusion} related to vision-language-action models~\cite{brohan2023rt2,kim2024openvla,black2024pi0}; the environment is agnostic to the policy.

\section{Methods}

\subsection{Specimen-Specific Surgical Scene Reconstruction}

\begin{figure*}[t]
  \centering
  \includegraphics[width=\textwidth]{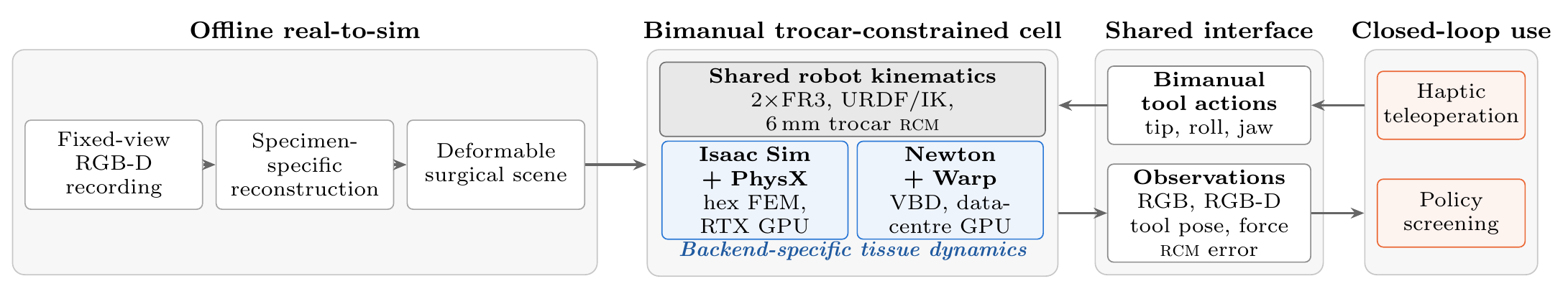}
  \caption{System overview. Both physics backends share robot kinematics and one action-observation interface; tissue dynamics are backend-specific.}
\end{figure*}

\begin{figure}[t]
  \centering
  \includegraphics[width=\columnwidth]{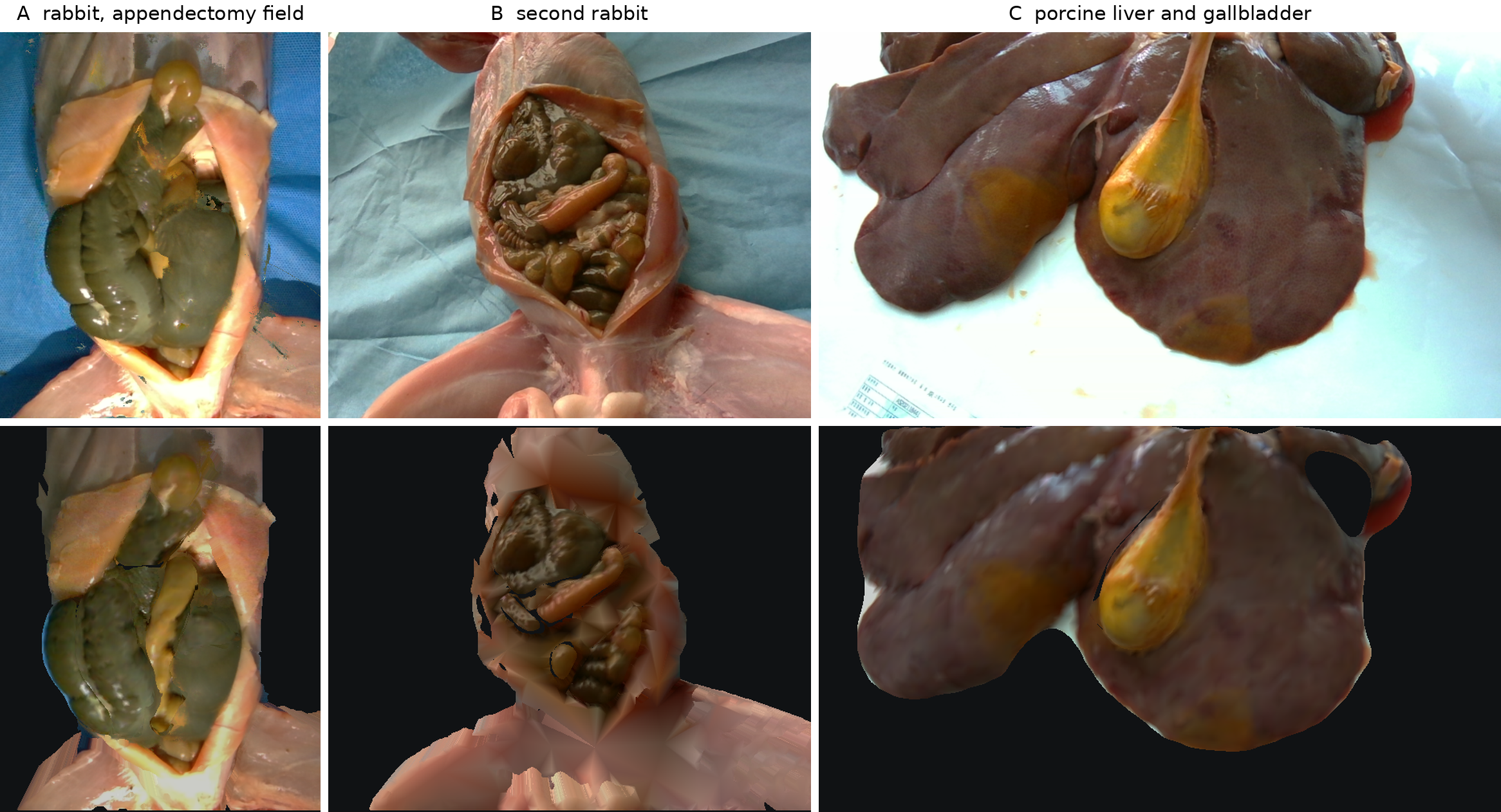}
  \caption{Reconstruction of three specimens. Top: recorded frames (A: instruments removed by compositing). Bottom: reconstructed surfaces rendered from the recording camera. Only specimen A is used in the robotic cell.}
\end{figure}

Each scene starts from one fixed-viewpoint RGB-D recording acquired during ex-vivo collection with the camera mounted vertically over the field. For the specimen used in the cell, the recording comprises 60~s of RealSense D405 data at 848$\times$480 and 30~fps (1789 colour and 1792 depth frames). Because the camera is fixed, pixel $(u,v)$ observes the same ray in every frame and frames are combined by indexing, without registration.

\textbf{Instrument removal.} Let $\mathcal{V}(u,v)$ be the frames in a window in which the pixel is neither masked as instrument nor invalid. The background is the per-pixel median, $\widehat{D} = \mathrm{median}_{\mathcal{V}}D_t$, and likewise for colour; pixels with $|\mathcal{V}|<3$ are left empty. A $\pm$90-frame window at stride~2 recovered 3.4\% of the image under the instrument mask and raised usable depth coverage to 90.3\%, at a median inter-percentile depth range of 2.0~mm. Organs are segmented with SAM~2.1~\cite{ravi2024sam2} from point prompts; instrument masks come from an appearance detector.

\textbf{Closing one observed surface into a volume.} FEM requires a closed volume, whereas a single viewpoint measures only the front surface. Missing depth inside a mask is filled by mask-confined Laplace diffusion. The back surface is then modelled: with $d$ the metric distance from an interior pixel to the silhouette and $R=P_{99.5}(d)$, the thickness is
\begin{equation}
  h = 2\sqrt{d\,(2R-d)},\quad z_{\mathrm{back}} = z_{\mathrm{front}} + h,
\end{equation}
so that the organ closes with a rounded profile. Occupancy is marked along $[z_{\mathrm{front}}, z_{\mathrm{back}}]$ on a 1.5~mm grid, morphologically closed, surfaced by marching cubes~\cite{lorensen1987marching} at the 0.5 level, reduced to its largest connected component, Taubin-smoothed~\cite{taubin1995signal} ($\lambda=0.5$, $\mu=-0.53$), and vertex-coloured by projection into the photograph. Watertightness is verified before further processing.

\textbf{Tetrahedralisation.} Each surface is tetrahedralised with TetGen~\cite{si2015tetgen} at a minimum dihedral angle of 10$^\circ$ and a radius-edge ratio of 1.6, and the signed volume $V = \frac{1}{6}[(\mathbf{b}-\mathbf{a})\times(\mathbf{c}-\mathbf{a})]\cdot(\mathbf{d}-\mathbf{a})$ of every element is checked; inverted elements ($V\leq 0$) are reoriented. Specimen~A yields six organs (cecum, three small-bowel segments, stomach and appendix) with 11,962 nodes and 43,956 tetrahedra (Table~\ref{tab:fem}), all watertight with no inverted elements. The median element edge length is $\approx$1.6~mm.

\begin{table}[t]
\caption{Reconstructed FEM Assets}
\label{tab:fem}
\centering
\begin{tabular}{llrrr}
\toprule
Scene & Organ & Nodes & Tets & $E$ (kPa)\\
\midrule
A: rabbit, robotic cell & Cecum        & 3,893  & 14,914 & 30\\
                        & Small bowel 1& 2,242  &  8,211 & 30\\
                        & Small bowel 2&   853  &  3,024 & 15\\
                        & Small bowel 3& 1,744  &  6,328 & 15\\
                        & Stomach      &   967  &  3,407 & 30\\
                        & Appendix     & 2,263  &  8,072 & 12\\
B: second rabbit        & Cecum        &   448  &  1,372 & 20\\
                        & Small bowel  & 1,339  &  4,878 & 20\\
                        & Bowel bloc   & 4,603  & 17,780 & 20\\
                        & Stomach      &   292  &    895 & 20\\
                        & Appendix     & 1,211  &  4,299 & 12\\
C: porcine liver        & Liver        &13,039  & 50,834 & 10\\
                        & Gallbladder  & 5,626  & 22,047 &  5\\
\midrule
Total, 13 organs        &              &38,520  &146,061 &   \\
\bottomrule
\end{tabular}\\
{\small All organs are watertight with no inverted tetrahedra. Moduli are literature values and are not identified from a recording.}
\end{table}

\subsection{Deformable Tissue Modeling}

\subsubsection{Constitutive Model}

Each organ is an isotropic hyperelastic solid discretised over its tetrahedra. With Young's modulus $E$ and Poisson ratio $\nu$, the Lam\'{e} parameters are $\mu=E/(2(1+\nu))$ and $\lambda=E\nu/((1+\nu)(1-2\nu))$. With $\mathbf{F}$ the deformation gradient of a linear element and $\mathbf{E}=\frac{1}{2}(\mathbf{F}^\top\mathbf{F}-\mathbf{I})$ the Green strain, the energy density is
\begin{equation}
  \Psi(\mathbf{F}) = \mu\,\mathrm{tr}(\mathbf{E}^2) + \frac{\lambda}{2}\,\mathrm{tr}(\mathbf{E})^2,
\end{equation}
integrated per element to give nodal forces $\mathbf{f}=-\partial(\sum_e V_e\Psi_e)/\partial\mathbf{x}$, with $\nu=0.45$ and velocity damping on the element.

Moduli are literature values~\cite{rosen2008biomechanical} chosen per organ. The stand-alone VBD scenes of specimens B and C use $E=20$~kPa for rabbit organs with the appendix at 12~kPa, 10~kPa for porcine liver and 5~kPa for the thin-walled gallbladder, at $\rho=1050$~kg\,m$^{-3}$. The bimanual cell uses 30~kPa for the cecum, first small-bowel segment and stomach, 15~kPa for the two thinner bowel segments, 12~kPa for the appendix and $\rho=1200$~kg\,m$^{-3}$.

\subsubsection{Implicit Integration with Vertex Block Descent}

At an element size of $\approx$1.6~mm, a stable explicit step would require $\Delta t\sim 10^{-5}$~s. The two non-implicit alternatives tested, semi-implicit integration and XPBD~\cite{macklin2016xpbd}, reach the particle velocity limit of the scene on the first frame and remain there (\S{}VII-B); this did not change with substep count or with a uniform mass distribution, which indicates that element size rather than mass ratio is the cause. We therefore integrate with Vertex Block Descent (VBD)~\cite{chen2024vbd}, which recasts one implicit Euler step as
\begin{equation}
  \mathbf{x}_{n+1} = \arg\min_{\mathbf{x}}\;\frac{1}{2\Delta t^2}\|\mathbf{x}-\mathbf{y}\|_{\mathbf{M}}^2 + E(\mathbf{x}),
\end{equation}
with $\mathbf{M}$ the lumped mass matrix, $E$ the elastic and contact potential and $\mathbf{y}=\mathbf{x}_n+\Delta t\,\mathbf{v}_n+\Delta t^2\mathbf{M}^{-1}\mathbf{f}_{\mathrm{ext}}$ the inertial predictor. VBD minimises~(3) by block coordinate descent, each vertex taking a local Newton step on its $3\times3$ block, with vertices graph-coloured so that blocks of the same colour update in parallel. We use $\Delta t=1/60$~s with 8 substeps of 10 iterations; free fall matches the analytic solution to four significant digits.

\subsubsection{Discretization and Mesh Separation}

The PhysX backend cooks a hexahedral simulation mesh, which avoids the sliver elements that made every tetrahedral path on this data unstable. Voxelisation, however, adds stiffness unrelated to the material: at the automatic resolution the appendix received 154 nodes and, spanned by a few large hexahedra, could not drape over the jaws at any modulus. The resolution is therefore fixed at 16 cells along the longest axis of each organ (8,065 nodes for six organs); $E$ does not describe tissue behaviour without this resolution.

\textbf{Separate Simulation and Collision Meshes.}
A hexahedral simulation mesh cannot serve as the collision mesh: cooking fails for the whole body without an error, and the organ renders and is skinned but does not move. For an isolated cube over 1~s, the shared-mesh configuration moved 0.00~mm while every other configuration fell $\approx$4.9~m. Each organ is therefore represented by a cooked hexahedral simulation mesh, a cooked tetrahedral collision mesh and a render mesh skinned by PhysX.

\subsubsection{Numerical Guards and Boundary Conditions}

\textbf{Mass floor.} Delaunay refinement leaves sliver elements whose nodes receive $\sim\!10^{-9}$~kg against a median of $\sim\!10^{-4}$~kg. Masses are set to $m_i\leftarrow\max(m_i,\,0.25\,\mathrm{median}_j\,m_j)$, which affects 8.4\% of nodes on specimen~A; under VBD this guard did not change the result (\S{}VII-B) and is retained as a precaution.

\textbf{Pinning.} Mesentery and ligaments are not observed, so the lower band of nodes of each organ is held at its captured position and the appendix is additionally fixed at the end farthest from the grasping arm ($\approx$42\% of nodes). Fixing must be applied via the builder before finalisation; zeroing inverse mass afterwards left nodes free (38.6~mm drift in 1.5~s).

\textbf{Support geometry.} Organ-field collision pushes each organ out of its own imprint (the stomach moved 24~mm in one second), so this pair is filtered and a conforming surface supports the appendix.

\subsection{Contact and Grasp Modeling}

\subsubsection{Tool--Tissue Contact}

With a near-zero collision offset, packed organs register contact only after overlapping, so the cell uses a 1.5~mm contact offset, zero rest offset, friction 0.6 and 48 solver iterations; instrument tips carry a 1.5~mm rest offset.

\subsubsection{Live-Pose Grasp Constraint}

Friction is insufficient for tissue of a few kilopascals: with the grasp assist disabled, 0 of 15 repeated grasp-lifts held the appendix (\S{}VII-D), and raising the modulus until friction holds would misrepresent the organ in all other interactions. The grasp is therefore a kinematic constraint on a subset of tissue nodes, defined by three choices.

\textbf{Where a grasp may form.} The free part of the appendix is partitioned into disjoint clusters of approximately 1~cm along its axis, so that a cluster index is a position on the organ; this makes a grasp-location score expressible (\S{}VI-C). The jaw mouth is the measured blade geometry, a wedge from 2 to 21~mm beyond the pivot and 5~mm wide, whose half-gap follows $r\sin(\theta/2)$. Tissue is considered grasped only if it lies in the wedge on both sides of the closing plane and the blades are farther apart than their contact skin; the attached cluster is the one nearest the centroid of that tissue, within 12~mm.

\textbf{Activation pose.} Let $t_a$ be the step at which the jaws close on in-mouth tissue, $(\mathbf{R}(t),\mathbf{p}(t))$ the jaw frame and $\mathbf{x}_i$ the cluster nodes. Offsets are captured from the live tissue at activation,
\begin{equation}
  \mathbf{o}_i = \mathbf{R}(t_a)^\top\!\bigl(\mathbf{x}_i(t_a)-\mathbf{p}(t_a)\bigr),
\end{equation}
and each subsequent step prescribes $\mathbf{x}_i(t)=\mathbf{R}(t)\,\mathbf{o}_i+\mathbf{p}(t)$ with $\mathbf{v}_i=\mathbf{0}$ until the jaws open. The constraint residual at activation is zero by construction.

\textbf{Effect of the capture pose.} An attachment authored at the rest pose welds nodes at offsets $\mathbf{o}_i^{\mathrm{rest}}$ and, when enabled at $t_a$, imposes in one step
\begin{equation}
  \boldsymbol{\Delta}_i = \mathbf{R}(t_a)\,\mathbf{o}_i^{\mathrm{rest}} + \mathbf{p}(t_a) - \mathbf{x}_i(t_a),
\end{equation}
which equals the displacement of the organ since rest. After a scripted approach, $|\boldsymbol{\Delta}|\approx22$~mm, and the resulting FEM reaction displaced the instrument 13~mm out of its 6~mm port. Live capture~(4) sets $\boldsymbol{\Delta}_i=\mathbf{0}$; its effect is measured in \S{}VII-D.

\subsubsection{Contact-Force Estimation}

At the tool mount, the 4.7~N instrument weight exceeds tissue contact forces of tens of millinewtons, so the PhysX backend identifies a wrench model over 54 static poses and tares at reset. The Newton backend reads force directly from the soft-contact penalty model. During a latched grasp the channel reports constraint force rather than tissue reaction.

\subsection{Bimanual Trocar-Constrained Surgical Cell}

\subsubsection{Robot and Trocar Configuration}

Two FR3 bases stand 0.900~m apart with a 0.46~m stand-off from the phantom (Fig.~4). The left arm carries a motorised grasper and the right a motorised cutter, each through a 6~mm port. Because the shaft must pass through the port, tip pitch and yaw are determined by tip position; each arm has four commandable degrees of freedom (insertion, two pivots and roll) and the jaw, and the interface returns any rejected rotation to the caller.

A 5~mm shaft in a 6~mm hole has 0.5~mm of lateral clearance at normal incidence, decreasing to zero at $\approx$34$^\circ$ (Fig.~5). At the start pose the arms are tilted by 13.3$^\circ$ and 18.3$^\circ$, leaving 0.43 and 0.37~mm. The controller, selected by a 182-configuration grid search, holds the $p_{95}$ deviation at 0.34~mm and tracks free motion with 0.048~mm RMS error and a 0.26~mm static droop.

\begin{figure}[t]
  \centering
  \includegraphics[width=\columnwidth]{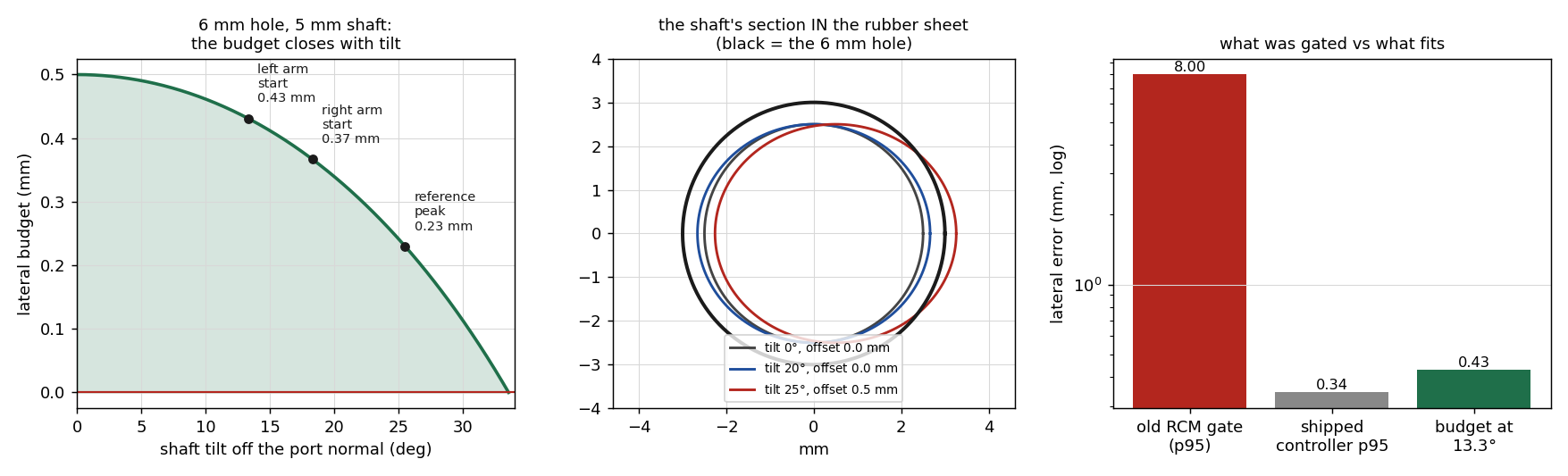}
  \caption{Port budget. Left: lateral clearance against shaft tilt with the start poses. Centre: shaft section in the port sheet. Right: admissible deviation.}
\end{figure}

\subsubsection{Camera Models and Simulation Backends}

The endoscope is a 30$^\circ$ forward-oblique laparoscope with the intrinsics of the real tower (ChArUco, 30 frames, RMS 0.58~px), undistorted to an 86.2$^\circ$$\times$54.6$^\circ$ pinhole at 1280$\times$720. The D405 model is placed above the port midpoint; the pose of the recording camera is recovered from the transform that places the reconstruction in the cell (\S{}VII-F). Two backends run under one contract (Fig.~2): arm motion is identical, and tissue deformation may differ. Backend~A is Isaac Sim~5.1~\cite{nvidia2025isaac} with PhysX FEM on an RTX~4090; backend~B is Newton (Warp~\cite{macklin2022warp}) with VBD and a path tracer that requires no RT cores, for batched evaluation. The second backend parses the cell constants from the build script of the first.

\subsubsection{Task Definition and Scoring}

The task is an ordered five-stage resection (grasp, lift, positioning, jaw on tissue, cut) with both instruments inside their ports, scored with the protocol of the companion policy study~\cite{anonymous2026bimanual}. The grasp must land at a normalised position of 0.30 to 0.70 along the appendix, and the cut must open the blades past 0.50 and close them below 0.15 through tissue; the organ is not topologically severed. Each episode also reports RCM deviation against a 3~mm threshold, lift, tracking lag and peak force.

\section{Experiments}

\begin{figure*}[t]
  \centering
  \includegraphics[width=\textwidth]{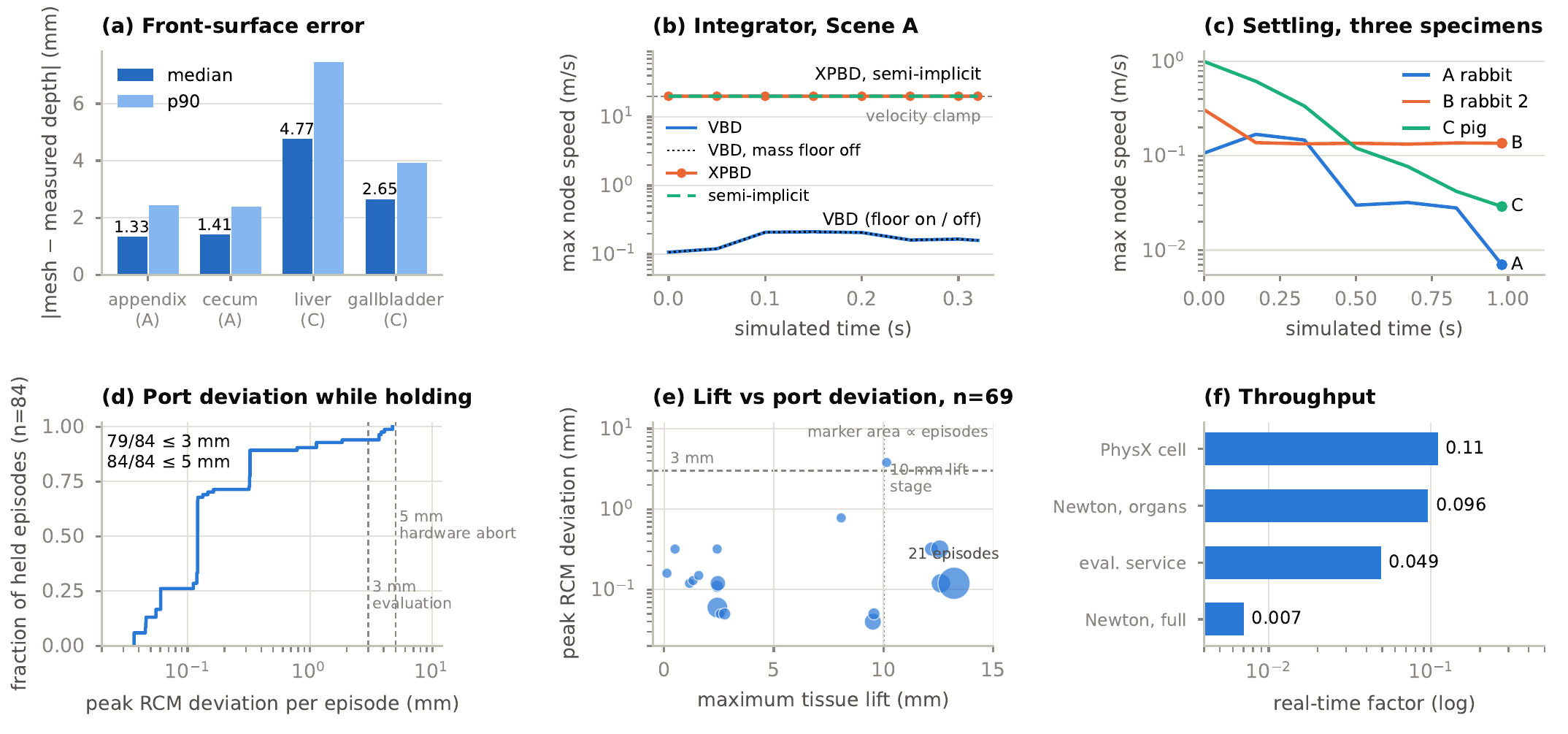}
  \caption{Quantitative behaviour. (a)~Front-surface error. (b)~Integrator comparison, specimen~A. (c)~Settling of the three specimens. (d)~Peak RCM deviation in 84 closed-loop episodes with tissue held. (e)~Lift against peak RCM deviation. (f)~Real-time factor by backend.}
\end{figure*}

\subsection{Reconstruction Fidelity}

Each organ mesh is rasterised through the pinhole model of the recording and its front surface is compared pixel by pixel with the depth map from which it was built (Fig.~6(a)); the back surface, modelled by~(1), is excluded. On the rabbit scene the median error is 1.33 to 1.41~mm over 69 to 92\% of the organ pixels, close to the 1~mm depth quantisation plus half a voxel. The porcine liver, the largest and flattest structure, has a larger error of 4.77~mm. The signed error (mesh minus measured depth) is negative on all four organs, from $-$1.15 to $-$4.88~mm, consistent with marching cubes placing the surface half a cell beyond the occupied cells and with morphological closing filling concavities outward.

\subsection{Numerical Stability}

On specimen~A at 60~fps, semi-implicit integration and XPBD reach the 20~m/s particle velocity limit on the first frame and remain there, so this value bounds their divergence rather than measuring it; VBD remains at 0.16~m/s (Fig.~6(b)). Disabling the mass floor leaves the maximum node speed unchanged to three decimal places. Released into the cell, the organs of specimen~A peak at 0.17~m/s and decay to 0.007~m/s within one second, without divergence or tunnelling.

\subsection{Cross-Specimen Reconstruction Robustness}

The reconstruction, tetrahedralisation and solver path was applied unchanged to a second rabbit from a different session and to a porcine liver with its gallbladder (Fig.~3, Table~\ref{tab:fem}). All thirteen organs are watertight with no inverted tetrahedra, organ volume spans 0.59 to 815~cm$^3$ (a factor of 1400) under the same closure model and TetGen settings, and all three scenes step without divergence (Fig.~6(c)). With 41.6\% and 6.6\% of nodes fixed, specimens~A and C settle to 0.007 and 0.029~m/s within one second, whereas specimen~B, with 21.1\% fixed, remains at 0.136~m/s.

\subsection{Contact and Grasp Stability}

\begin{table}[t]
\caption{Grasp Design Ablation (Newton cell, 15\,mm jaw lift)}
\label{tab:grasp}
\centering
\begin{tabular}{lccrrr}
\toprule
Design & Held & Diverged & Lift & Slip & Snap\\
       &      &          & (mm) & (mm) & (mm)\\
\midrule
Friction           & 0/15  & 0/15 &  4.9 & 10.1 & --\\
Rest-pose          & 13/15 & 1/15 & 15.1 &  0.45 & 4.7\\
Live-pose          & 13/15 & 2/15 & 15.2 &  0.32 & 0\\
Re-grasp, rest-pose& 5/5   & 0/5  & 15.4 &  0.48 & 4.3--17.8\\
Re-grasp, live-pose& 5/5   & 0/5  & 15.2 &  0.37 & 0\\
\bottomrule
\end{tabular}\\
{\small Medians over five grip locations and three approach offsets. Held: tissue follows at least 80\% of the jaw lift with less than 3\,mm slip.}
\end{table}

\begin{figure*}[t]
  \centering
  \includegraphics[width=\textwidth]{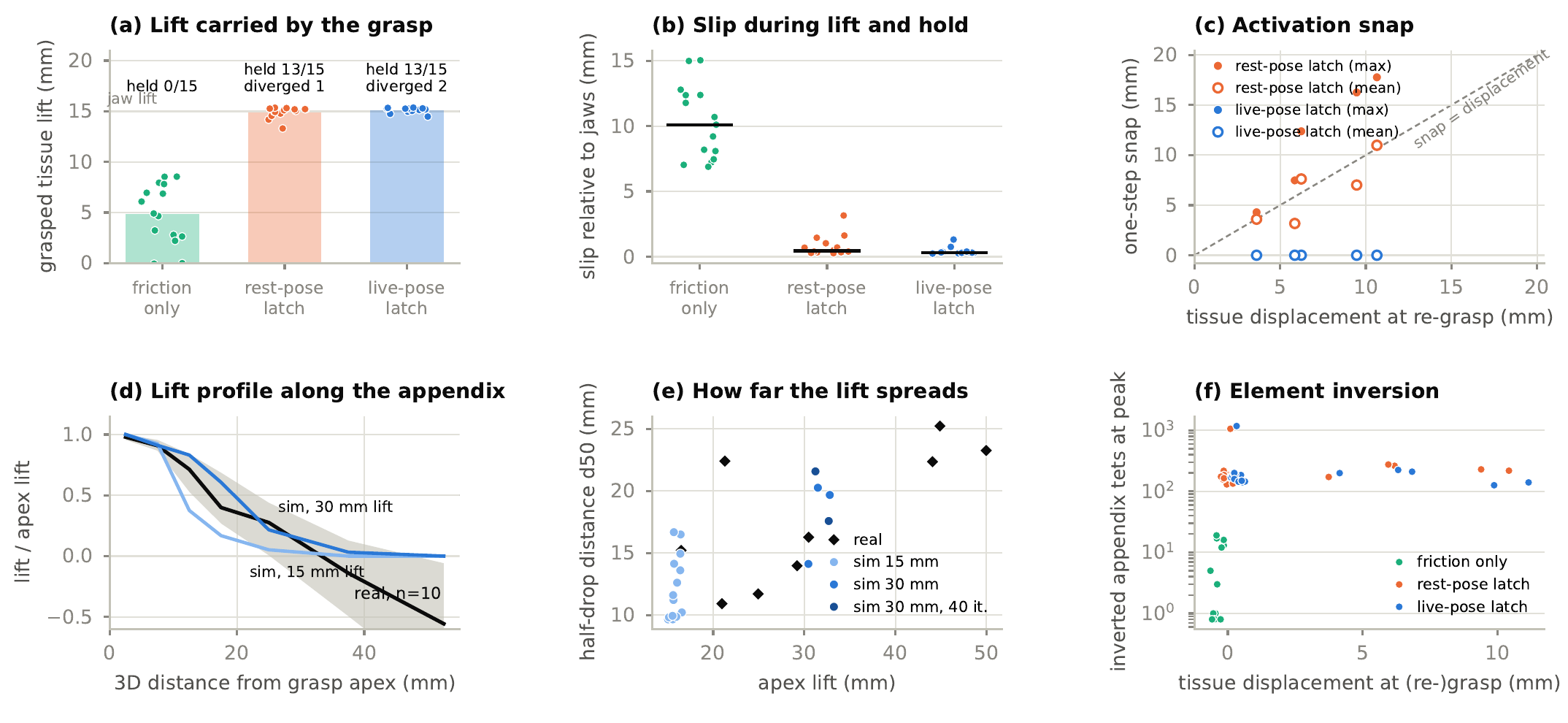}
  \caption{Grasp validation. (a)~Tissue lift per grasp design. (b)~Slip. (c)~Activation snap on re-grasp. (d)~Normalised lift profile, real and simulated. (e)~Half-drop distance against apex lift. (f)~Inverted appendix tetrahedra.}
\end{figure*}

\textbf{Repeated grasp-design ablation.} Each design was applied in the same scripted grasp-lift at five grip locations along the appendix and three lateral approach offsets (Table~\ref{tab:grasp}, Fig.~7(a,b)). The friction grasp held in no trial: the tissue rose 4.9~mm for 15~mm of jaw lift (median follow ratio 0.33) and slipped 10.1~mm. Both constraint grasps carried the tissue (follow ratio 1.01) in 13 of 15 trials; the remaining trials diverged, as discussed below.

\textbf{Activation on displaced tissue.} To isolate~(5), the jaws grasped and dragged the tissue by 5 or 10~mm, released it and closed again, with the second activation performed with rest-pose or live-pose capture from an otherwise identical state. At re-grasp the tissue was displaced by 3.6 to 10.6~mm from rest; rest-pose attachment produced one-step snaps of 4.3 to 17.8~mm (maximum over nodes), with the node-mean snap proportional to displacement (slope 0.91, $r=0.83$), whereas live-pose capture produced no such snap (Fig.~7(c)). The Newton instrument is kinematic, so the snap cannot move the tool there; in the articulated PhysX cell the same mechanism raised the RCM error to 16.4~mm with 6 of 60 control steps beyond 3~mm, and to 0 of 60 after the live-pose change.

\textbf{Element quality.} Carrying a cluster rigidly against the fixed far end inverts 170 to 180 of the 8,072 appendix tetrahedra at peak lift (1 with friction), and 3 of 30 constrained 15~mm lifts diverged (Fig.~7(f)). At 30~mm lift, divergence did not depend on solver iterations but on appendix modulus: 3/3 at 3 and 6~kPa, 1/3 at 12 and 24~kPa and 0/3 at 48~kPa.

\textbf{Holding under closed-loop load.} The jaw-mouth predicate was correct in all 15 placements of a synthetic tube. In 140 rollouts from six controllers, 84 episodes attached the tissue and 83 held it to the end; the median lift was 12.2~mm and peak RCM deviation had a median of 0.12~mm, within 3~mm in 79 episodes and within the 5~mm hardware abort threshold in all 84 (Fig.~6(d,e)). A scripted finite-state baseline passes all five stages (grasp at 0.50, lift 10.1~mm); when its grasp takes load the left RCM error rises to 2.6~mm, beyond the geometric clearance but inside the evaluation threshold, so port safety is reported separately from task progress.

\subsection{Comparison with Real Grasp-Lifts}

The recording session also contains the grasp-lifts performed by the surgeon. In 14 marked intervals the appendix was tracked with SAM~2 from one click; ten were tracked throughout and are used. The height of the appendix above a robust plane through the surrounding tissue gives an apex lift of 16.5 to 50.0~mm (median 29.8) and a half-drop distance $d_{50}$, at which lift falls to half, of 10.9 to 25.2~mm (median 16.3). $d_{50}$ increases with apex lift ($r=0.71$; $d_{50}\approx7.9+0.32\,h_{\mathrm{apex}}$ over nine intervals, residual SD 4.1~mm). Measured identically in simulation, the live-pose constraint gives $d_{50}=11.4$~mm at 15.7~mm apex lift and 19.7~mm at 31.5~mm, within 0.72 and 0.56 residual SD of the real trend (Fig.~7(d,e)). This agreement holds for appendix moduli of 3 to 48~kPa, so $d_{50}$ does not identify the modulus. Real apex lift is a lower bound, because the intervals begin with the appendix already grasped.

\subsection{Real-to-Sim Similarity}

\begin{figure*}[t]
  \centering
  \includegraphics[width=\textwidth]{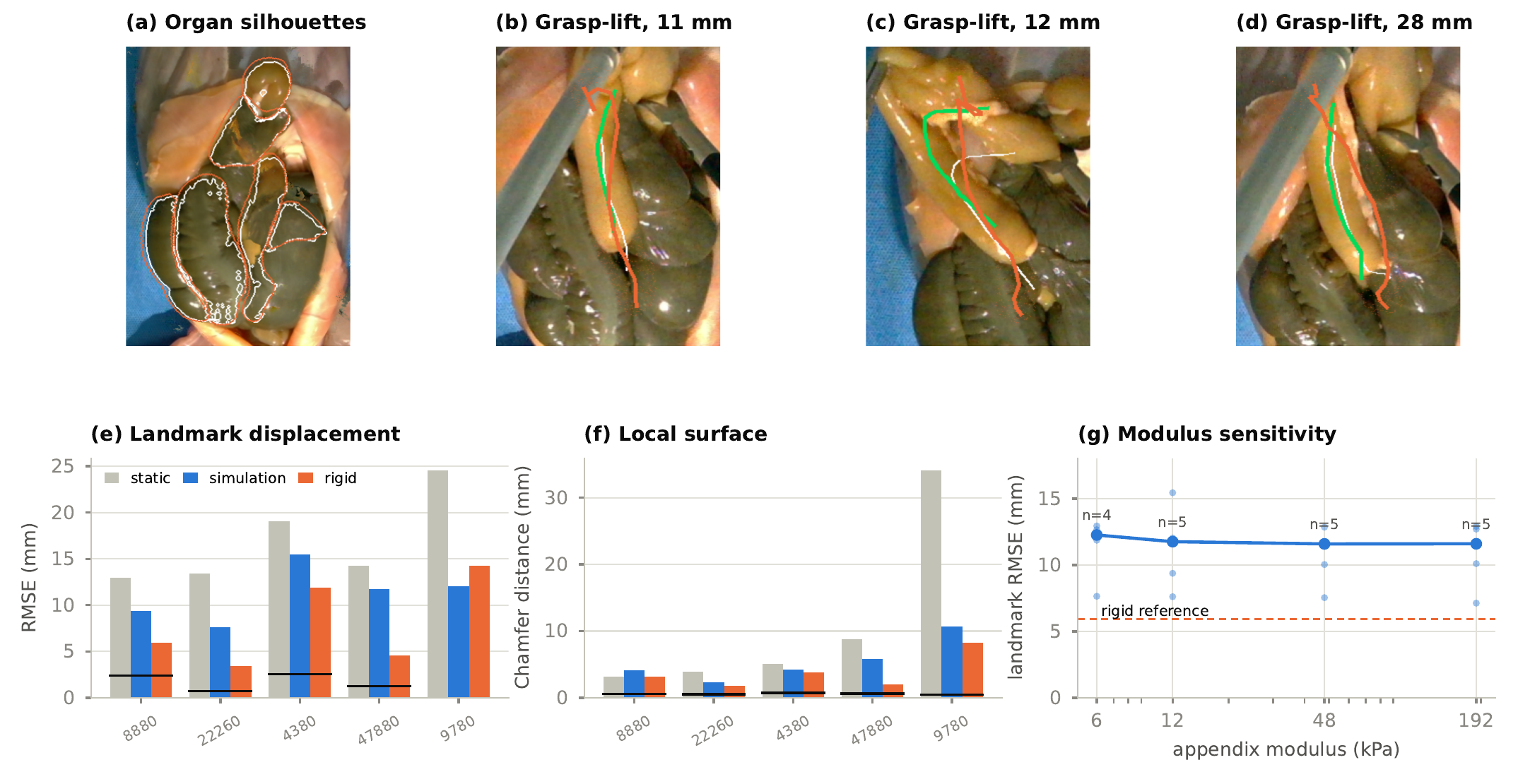}
  \caption{Real-to-sim similarity. (a)~Real (white) and simulated (orange) organ outlines. (b)--(d)~Matched grasp-lifts: real pre-grasp (white) and peak (green) centrelines, simulated centreline (orange). (e,~f)~Deformation error against static and rigid references; black lines mark the measurement floor. (g)~Sensitivity to appendix modulus.}
\end{figure*}

\begin{table}[t]
\caption{Real-to-Sim Similarity Metrics}
\label{tab:sim}
\centering
\begin{tabular}{lll}
\toprule
Metric & Simulation & Reference\\
\midrule
Depth error (mm)$^a$   & 1.33--1.41 & 1.0 quantisation\\
Silhouette IoU         & 0.80       & 0.24 cell camera\\
SSIM / PSNR (dB)$^b$  & 0.63 / 20.4& 0.67/19.5 real; 0.66/19.0 mean\\
Landmark RMSE (mm)     & 11.8       & 14.2 static; 5.9 rigid; 2.4 floor\\
Local Chamfer (mm)     & 4.3        & 5.1 static; 3.2 rigid; 0.6 floor\\
Force-displacement     & n/a$^c$    & --\\
\bottomrule
\end{tabular}\\
{\small Medians. $^a$Rabbit appendix and cecum. $^b$Exposure-matched render in organ masks; references: second real frame and mean-colour image. $^c$No force measurement in the recording.}
\end{table}

Similarity to the recording of specimen~A is summarised in Table~\ref{tab:sim}.

\subsubsection{Geometry and Appearance}

Projected through the pose of the recording camera, the simulated rest scene yields a mean organ silhouette IoU of 0.80 (range 0.76 to 0.83) (Fig.~8(a)); the cell D405 model, offset by 7.9~mm laterally and 4.6~mm vertically, gives 0.24. Inside the organ masks, SSIM~\cite{wang2004image} and PSNR of the path-traced view are 0.43 and 14.7~dB, rising to 0.63 and 20.4~dB after per-channel exposure matching; a second real frame and a mean-colour image score similarly (Table~\ref{tab:sim}), so SSIM has low discriminative power on this low-texture tissue.

\subsubsection{Deformation under Grasping}

Six real grasp-lifts with grasp-point displacements of 11 to 38~mm were reproduced in simulation: the pre-grasp centreline was rigidly aligned to the simulated rest centreline~\cite{besl1992method} (median residual 7.4~mm), the cluster at the same arc length was grasped and moved by the measured displacement; one reproduction diverged. Landmarks at 5~mm intervals and surface points within 25~mm of the grasp point were compared by RMSE and symmetric Chamfer distance~\cite{barrow1977parametric,fan2017point}, against a static prediction, a rigid prediction in which all landmarks follow the grasp point, and a floor from two real frames at the peak. The simulated landmark displacement RMSE was 11.8~mm (range 7.6 to 15.4), below the static prediction (14.2~mm) and above the rigid prediction (5.9~mm), and the local Chamfer distance followed the same order (4.3, 5.1 and 3.2~mm) (Fig.~8(e,f)). The simulated appendix therefore deforms more locally around the grasp than the real appendix. With the appendix modulus at 6, 48 and 192~kPa, the median RMSE changed only from 12.3 to 11.6~mm (Fig.~8(g)), so the discrepancy is not attributable to stiffness; the unobserved distal fixation and support of the appendix are the remaining candidates.

\subsection{Closed-Loop Screening and Throughput}

Neither backend runs in real time (Fig.~6(f)). The PhysX cell on an RTX~4090 advances an eight-step, 30~Hz action chunk in $\approx$2.4~s with physics at 480~Hz, of which 64 to 72\% is spent in substepping. On the H100, Newton runs the organs alone at 0.096$\times$ real time and a full episode with arms and rendering at 0.007$\times$, one environment per process.

The environment served as the closed-loop screening stage of a companion policy study~\cite{anonymous2026bimanual}, which selected its deployed configuration on port-safety and stage-progress counts in simulation; the protocol and hardware outcomes are reported there, and no rank correlation is claimed here. Binary stage gates required calibration: only 5 of 522 demonstrations closed the cutter below the 0.15 threshold, and rescoring 576 rollouts at thresholds of 0.15 to 0.50 yielded 0 to 62 passes, so continuous channels are reported alongside gate sweeps. Simulated teleoperation data (1,009 episodes) did not improve real-rig performance when mixed into training; a crossed ablation attributed most of the gap to appearance. The environment is therefore used for screening rather than for photorealistic training data.

\subsection{Extension to In-Vivo Surgical Scenes}

The route requires only a fixed-view capture and a segmentation. A scene built for an in-vivo laparoscopic session renders through the endoscope model beside the real view (Fig.~9); the result is a contact-stable, simulable copy of the observed specimen rather than an identified biomechanical model.

\section{Discussion and Limitations}

\subsection{Limitations of the Grasp and Boundary Models}

\textbf{No material identification.} Moduli are literature values (\S{}IV-A), and neither lift spread nor landmark displacement from video identifies them; force-displacement agreement requires force measurement, and 1~mm depth quantisation limits identification from small presses.

\textbf{Scope of the cell and of the grasp.} Only specimen~A is used in the robot cell. The grasp is a kinematic constraint that does not model slip, tearing or jaw pressure, inverts approximately 2\% of appendix elements at peak lift, and diverges in 30~mm lifts unless the appendix modulus is sufficiently high. Real-to-sim deformation was evaluated on five surgeon-performed grasp-lifts rather than robot-executed trials, and the simulated appendix deforms more locally than the real one. The environment is neither real-time nor run-to-run deterministic.

\section{Conclusion}

Specimen-specific reconstruction converts a fixed-view surgical RGB-D recording into watertight deformable anatomy, demonstrated on three specimens. Implicit integration, discretisation-aware modelling, separate collision and simulation meshes, numerical guards and a live-pose grasp constraint make bimanual manipulation through 6~mm trocars stable. Validation against measurement covered reconstruction error, a repeated grasp-design ablation, real-to-sim similarity (silhouette IoU 0.80; landmark RMSE 11.8~mm) and RCM behaviour in closed-loop rollouts. Biofidelity is not claimed.


\end{document}